\documentclass[10pt,twocolumn,letterpaper]{article}

\usepackage{cvpr}              % To produce the CAMERA-READY version
\usepackage{algorithm}
\usepackage{algpseudocode}
\usepackage{mathtools}
\usepackage{float}
\usepackage[accsupp]{axessibility}
\usepackage[dvipsnames]{xcolor}

\definecolor{cvprblue}{rgb}{0.21,0.49,0.74}
\usepackage[pagebackref,breaklinks,colorlinks,citecolor=cvprblue]{hyperref}

\def\paperID{49} % *** Enter the Paper ID here
\def\confName{CVPR}
\def\confYear{2024}

\title{ConPro: Learning Severity Representation for Medical Images \\using Contrastive Learning and Preference Optimization}

\author{
Hong Nguyen\textsuperscript{1} \and
Hoang Nguyen\textsuperscript{2} \and
Melinda Chang\textsuperscript{1} \and
Hieu Pham\textsuperscript{2,3} \and 
Shrikanth Narayanan\textsuperscript{1} \and
Michael Pazzani\textsuperscript{1} \and
\textsuperscript{1}University of Southern California, Los Angeles, United States\\
\textsuperscript{2}College of Engineering \& Computer Science, VinUniversity, Hanoi, Vietnam\\ 
\textsuperscript{3}VinUni-Illinois Smart Health Center, VinUniversity, Hanoi, Vietnam\\
{\tt\small \{hongn, shri, mpazzani\}@usc.edu,
\tt\small {mchang@chls.usc.edu}, 
\tt\small {hieu.ph@vinuni.edu.vn}}
}

\begin{document}
\maketitle
\begin{abstract}
Understanding the severity of conditions shown in images in medical diagnosis is crucial, serving as a key guide for clinical assessment, treatment, as well as evaluating longitudinal progression. This paper proposes \textbf{ConPrO}: a novel representation learning method for severity assessment in medical images using \textbf{Co}ntrastive learning-integrated \textbf{Pr}eference \textbf{O}ptimization. 
Different from conventional contrastive learning methods that maximize the distance between classes, ConPrO injects into the latent vector the distance preference knowledge between various severity classes and the normal class. 
We systematically examine the key components of our framework to illuminate how contrastive prediction tasks acquire valuable representations.
We show that our representation learning framework offers valuable severity ordering in the feature space while outperforming previous state-of-the-art methods on classification tasks. We achieve a 6\% and 20\% relative improvement compared to a supervised and a self-supervised baseline, respectively. In addition, we derived discussions on severity indicators and related applications of preference comparison in the medical domain. \textit{Code available at https://github.com/Hong7Cong/ConPrO.git}
\end{abstract}
\section{Introduction}\label{sec:intro}
\begin{figure}[t]
  \centering
  \begin{subfigure}{0.48\linewidth}
    \includegraphics[width=0.9\linewidth]{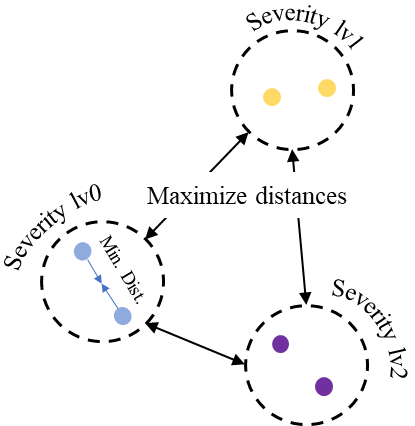}
    \caption{Conventional}
    \label{fig:ConventionalCL}
  \end{subfigure}
  \hfill
  \begin{subfigure}{0.48\linewidth}
    % \fbox{\rule{0pt}{2in} \rule{.9\linewidth}{0pt}}
    \includegraphics[width=\linewidth]{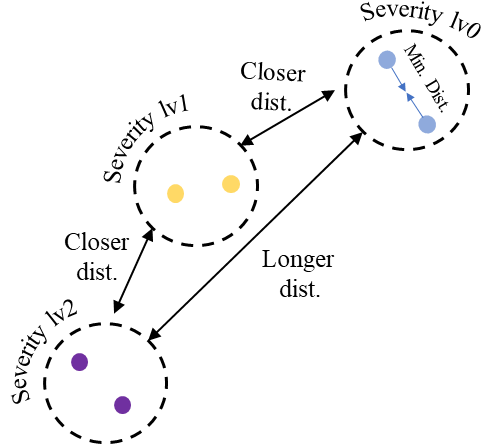}
    \caption{Proposed ConPro}
    \label{fig:OurCL}
  \end{subfigure}
  \caption{Conventional (a) and target (b) representation for severity modeling in latent space. A darker color represents a higher severity level, and `0' represents normality. Our proposed method targets to embed distance relation to severity classes in representation space}
  \label{fig:short}
\end{figure}

Recent advances in supervised \cite{supcon20, maxmarginCL22, Salakhutdinov2007LearningAN, NIPS2005_a7f592ce} and self-supervised contrastive learning \cite{simmim20, simclr21, Hjelm2018LearningDR, Zhai2019S4LSS} offer a strong foundation for image understanding and interpretation, including in medical applications. 
Crucially, latent vectors are acquired from data to capture increasing amounts of contextual information within an image and across contextual classes. 
Self-supervised contrastive learning attempts to exploit domain knowledge by bringing `positive' samples closer together in the embedding space while pushing `negative' samples apart. A positive pair often consists of augmented versions of the same sample and negative pairs are created using the anchor and randomly selected samples from the data batch.
On the other hand, supervised contrastive learning (SupCon) studies cross-class relations by grouping embeddings from the same class to the same cluster and pushing different clusters far from each other.

\begin{figure*}[t]
  \centering
  \begin{subfigure}{0.74\linewidth}
    \includegraphics[width=\linewidth]{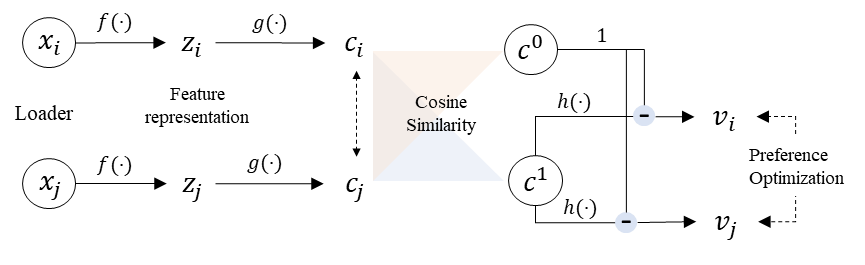}
    \caption{Learning Framework}
    \label{fig:framework}
  \end{subfigure}
  \hfill
  \begin{subfigure}{0.23\linewidth}
    % \fbox{\rule{0pt}{2in} \rule{.9\linewidth}{0pt}}
    \includegraphics[width=\linewidth]{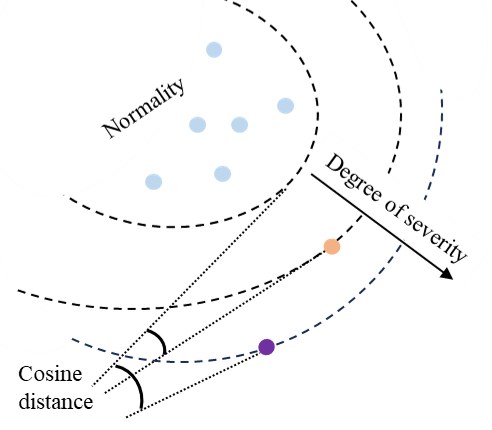}
    \caption{Desired latent space}
    \label{fig:short-b}
  \end{subfigure}
  \caption{ConPro learning framework (a) includes contrastive learning and preference optimization to get the desired latent space (b)}
  \label{fig:short}
\end{figure*}

% Conventionally, supervised contrastive learning treats all classes equally and independently, by maximizing the inter-class distance, as shown in Fig. ~\ref{fig:ConventionalCL}. However, classes may not always have an independent contextual relation. Due to differences in similarity between classes, some classes should be further away than others. For instance, class ``dog" should have a closer relation to ``wolf" than ``table."  In the medical domain, Fig. \ref{fig:OurCL}, the conditions of similar severity should have a smaller distance than those of a large severity difference.
Conventionally, supervised contrastive learning treats all classes equally and maximizes inter-class distance, as shown in Fig. \ref{fig:ConventionalCL}. However, this approach ignores the different level of similarity between classes, some classes should be further away than others. For instance, class ``dog" should have a closer relation to ``wolf" than ``table."  Similarly, in the medical domain (Fig. \ref{fig:OurCL}), the conditions of similar severity should have a smaller distance than those of large severity differences. Furthermore, bridging the gap from the non-medical images to the medical domain presents a distinct challenge since medical images are burdened by label sharpness \cite{konz2024the}, experts' annotation biases \cite{Petersen2019EuropeanAO, YANG2020109008}, weak labels \cite{weaklabel20, octweaklabel22}, and noise from various imaging modalities \cite{Morra2020BridgingTG}. These challenges may degrade the learned knowledge within the latent representation of medical images.

Despite playing an essential role in clinical practice, severity disparity in medical images has not been investigated well in the computer vision literature. 
% To this end, several works have learned severity ranking using the Siamese network. 
A common approach is to explore image-derived clinical severity as a multi-class classification problem \cite{SimSiamese2020, YILDIZ201965, diaRetino23, Qiblawey2021DetectionAS, xrayclassify22, hieu2023}, where each class corresponds to a different (quantized) severity level. The classifier then learns to distinguish attributes/traits that are probabilistically different between severity classes. However, a higher severity level may still share some traits or attributes with lower severity levels.
Thus, another proposed approach was to rank images with respect to severity scores.
In \cite{TianROC19}, the authors introduced the concept of severity ranking for fundus photography. Following this idea, \cite{nguyen2023explainable, YILDIZ201965, KalpathyCramer2016os, segmentandclassify19, Qiblawey2021DetectionAS} improved that work by incorporating multi-scoring, multitasking learning, and experts' agreement.
However, an undesirable outcome of current severity ranking methods is that all outputs can ``collapse" to a constant value. Thus, a single score for the severity of the entire image is a significant loss of information and interpretability. Our work addresses this challenge by introducing ranking loss in the representation space.
% Another approach is to explore image severity as multi-class classification problem \cite{SimSiamese2020} where each class correspond to different level of severity.
% The challenge raise in contrastive learning is that the models learn to distinguish attribute or traits of different classes. In severity evaluation, higher severity level may contain traits and attributes from lower levels.
% This behavior can degrade the scaling performance of classification.

% Inspired by previous works on representation learning, severity ranking we
In this paper, we inject severity information into a latent space vector using contrastive learning-integrated preference optimization (ConPrO). Not only does ConPrO show a reliable representation arrangement, but it also outperforms state-of-the-art (SOTA) algorithms on classification tasks. The contributions of this paper are summarized below:
\begin{itemize}
    \item We propose ConPrO: a novel representation learning method that incorporates class severity information within the latent space. ConPrO performs better in the F1 score, by a relative 20\% compared to SimCLR (self-supervised) and 6\% compared to SupCon (supervised).
    \item We introduce an evaluation metric, Mean Absolute Exponential Error (MAEE), for a specific problem (severity classification). MAEE penalizes incorrect prediction at higher severity classes in case of data imbalance.
    \item  We show that increasing the number of reference vectors helps reduce MAEE 
    % and suggest carefully choosing severity indicators in the continuous/regression case 
    and offer discussion on the potential application of preference comparison in the medical domain.
\end{itemize}

\section{Related Works}
\label{sec:formatting}
% This work raise a question on how to learn and interpret severity among medical images.

\noindent
\textbf{Visual Contrastive Learning.} The fundamental concept of contrastive learning is to push the latent vectors of different classes far apart from each other while pulling latent vectors within the same class closer. This method was introduced in representation learning as a self-supervised \cite{simmim20, simclr21, Hjelm2018LearningDR, Zhai2019S4LSS} or supervised way \cite{supcon20, maxmarginCL22, Salakhutdinov2007LearningAN, NIPS2005_a7f592ce} to inject relative information to embeddings in the latent space. Shekoofeh et al. \cite{remedis2023} show that contrastive learning improves the robustness and data efficiency of medical imaging tasks. Furthermore, the authors suggest that visual representation learning is a key component for building large (foundation) vision models.
% However, it is not the case for severity classes where nearby class-pair in the sequence of severity. For instance, in a 5-level papilledema severity rating, the 4-th and 5-th is closer than 4-th and 0-th. This challenge inspire us thinking for a approach to
Moreover, recent works \cite{Oord2018RepresentationLW, Hjelm2018LearningDR, predictivecoding20} have made efforts to relate the success of contrastive learning from the perspectives of mutual information, choices of feature encoder, and loss function. 
% Yet, previous methods not take relative distance between classes into account. As such, severity-levels dependency (Figure \ref{fig:OurCL}) bring up a new problem to representation learning.

% \noindent
% \textbf{Clustering} 

\noindent
\textbf{Preference comparison.} Originating from learning to rank problems \cite{ltr2005, Xia2019PlackettLuceMF} in recommendation systems, several works \cite{TianROC19, nguyen2023explainable} have used preference comparison for ranking disease severity. Yu \cite{RDRL20} proposed Relative Distance Ranking Loss, which measures the similarity between image patches and their reference image. Recently, preference comparison was re-introduced in Large Language Models (LLMs) as a way to optimize preference of generative pairs of answers given an input. RLHF \cite{NEURIPS2022_RLHF} learns the reward function from pairwise comparisons of output text and optimizes it via reinforcement learning. More recently, Direct Preference Optimization (DPO) \cite{DPO2023} simplifies RLHF by optimizing a language model directly to align with human preferences without relying on explicit reward modeling or reinforcement learning. DPO updates parameters by maximize preferences likelihood between pair of samples, which can be a potential approach toward severity ranking in latent spaces. 

%Within the author's knowledge, 
We note that none of the previous works investigate medical, image-driven preference with respect to severity, nor relative distance between classes. As such, severity-level dependency (Figure~\ref{fig:OurCL}) brings up a new problem in representation learning.
We also note that although the individual components of our framework have been presented in prior research,
% , although particular choices of function may be different to fit with our problem. 
the innovation of our framework is its combination to solve severity ranking problems with particular choices of reward/loss function.

\section{Method}
The framework, as shown in Fig.~\ref{fig:framework}, contains two main consecutive phases including binary contrastive learning (``Con'' step) and preference optimization (`P`rO'' step). In the `Con' step, we maximize the latent distance between the normal and the abnormal class. In the ``Pro'' phase, we re-arrange the relative distance of severity levels within abnormal classes with respect to reference vectors in the normal class. The detailed implementation and loss function of both steps are presented in the following.

\subsection{Contrastive Learning}

We group severity classes into a single abnormal class and perform binary contrastive learning between normal\footnote{The term ``normal'' in this paper refers to non-abnormal cases; e.g., images categorized as ``normal'' with respect to a specific pathology may not necessarily indicate healthy condition} and abnormal samples. 
The motivation for this phase is to group the positive samples in a cluster that is well-separated from the abnormal cluster in the latent space.
Subsequently, the normal cluster is used as an anchor for preference comparison. The framework contains a feature extractor $f(\cdot)$ which is a convolutional neural network that encodes images to latent vectors. The contrastive head $g(\cdot)$ maps those vectors to the 
contrastive space. 

\noindent
\textbf{Supervised Contrastive Objective.} For simplicity, we use margin contrastive loss (although there are multiple attractive options such as NT-Xent, XT-Logistic \cite{simclr21}) % that proved to be more robust)
\begin{equation}
\begin{aligned}
\mathcal{L}_\textit{Con}({c_i, c_j, y_{ij}} &) = \mathop{\mathbb{E}}
[
y_{ij} \, {d}_\text{cos}(c_i, c_j) + ...\\
&(1-y_{ij}) \, \textbf{max}(0, m-{d}_\text{cos}(c_i, c_j))
]
\label{equal:contrastive}
\end{aligned}
\end{equation}
where $m$ is maximum margin, ranging from 0 to 2 (here we choose $m=2$) and ${d}_\text{cos}$ denotes the cosine distance
\begin{equation}
    {d}_\text{cos}(c_i, c_j) = \frac{c_i^\top c_j}{\|c_i\| \, \|c_j\|}
\end{equation}
between a pair of vectors $(c_i, c_j)$.

\subsection{Preference Optimization over Latent Space}

Inspired by current advances in learning to rank algorithms as well as preference optimization algorithms in Natural Language Processing (NLP) such as RLHF \cite{NEURIPS2022_RLHF} and DPO~\cite{DPO2023}, our objective is to present a simple approach for severity comparison over the representation space.

\noindent
\textbf{Preference Comparison Objective.} Both RLHF and DPO use Bradley-Terry preference model \cite{bradley1952rank} to construct the loss function. Given some prior knowledge $\pi_0$, the Bradley-Terry model calculates the preference likelihood over a pair of samples with respect to labelers' severity measurement, denoted as $\nu_i > \nu_j | \pi_\text{0}$, where $\nu_i$ and $\nu_j$ are the preferred and dispreferred completion amongst ($\nu_i$, $\nu_i$) respectively. 
% The given severity preference distribution can be written as $p^*(\nu_i > \nu_j | \pi_\textbf{0}) = \sigma(r^*(\nu_i, \pi_\text{0})-r^*(\nu_j, \pi_\text{0}))$. 
The preferences are assumed to be generated by some predefined reward model $r^*$. 
In this work, we choose the reward function $r^* = {d}_\text{cos}$ as the cosine distance from severity to normality. Intuitively, we try to pull the less severe latent vectors closer to the ``normality'' anchor while pushing more severe cases far apart. We define the ``normality'' anchor as a vector or set of vectors belonging to the normal class. 
% We choose ``normality'' as an anchor to compare severities.
% The definition of the normal vector will be given and discussed later in this section. 
Under the Bradley-Terry model, we derive a simplified probability measure for pairwise severity comparison:
\begin{equation}
\begin{aligned}
  p^*(\nu_i > \nu_j | \pi_\textbf{0}) &= \sigma(r^*(\nu_i, \pi_\text{0})-r^*(\nu_j, \pi_\text{0}))\\
  &= \frac{1}{1 + \text{exp}[{d}_\text{cos}(\nu_i, \pi_0)-{d}_\text{cos}(\nu_j, \pi_0)]}
  \label{eq:important}
\end{aligned}
\end{equation}

where $x_\textbf{0}$ represented normality and $\text{d}_\text{cos}$ is the cosine distance. We can then formulate the problem in hand as binary classification and use the negative log-likelihood loss to re-parameterize the feature space:
\begin{equation}
\begin{aligned}
  \mathcal{L}_\textit{PrO}(\nu_i, \nu_j,& \pi_\text{0}, y_{ij}| r^*=d_\text{cos}) \\
  &= \mathbb{E}[\text{log}(\sigma(r^*(\nu_i, \pi_\text{0})-r^*(\nu_j, \pi_0)))]
  \label{eq:proloss}
\end{aligned}
\end{equation}

\begin{algorithm}[t]
\caption{ConPro Pseudocode}\label{alg:cap}
\begin{algorithmic}
% \State 
\Require Pre-defined $f,g,h$, contrastive loader $\mathcal{C}$, preference loader $\mathcal{P}$
\For{($x_i$, $x_j$, $y_{ij}$) in $\mathcal{C}$} \algorithmiccomment{\small \textit{`Con' step}}
\State $z_i$, $z_j \gets f(x_i)$,  $f(x_j)$
\State $c_i$, $c_j \gets g(z_i)$,  $g(z_j)$
% \State $l(i,j) = y_{ij} \, {d}_\text{cos}(c_i, c_j) + (1-y_{ij}) \, \textbf{max}(0, m-{d}_\text{cos}(c_i, c_j)$
\State Calculate $\mathcal{L}_\textit{Con}(c_i, c_j, y_{ij})$ 
\algorithmiccomment{Eq. (\ref{equal:contrastive})}
\State \text{Update network $f$ and $g$ to minimize} $\mathcal{L}_\textit{Con}$
\EndFor
\For{($c_i$, $c_j$, $\pi_\textbf{0}$, $y_{ij}$) in $\mathcal{P}$} \algorithmiccomment{\small \textit{`PrO' step}}
\State $\nu_i$, $\nu_j \gets h(c_i)$,  $h(c_j)$
\State Draw $\pi_0$ from $\mathcal{P}$
\State Calculate $\mathcal{L}_\textit{PrO}(\nu_i, \nu_j, \pi_\text{0}, y_{ij})$ \algorithmiccomment{Eq. (\ref{eq:proloss})}
% \State $\mathcal{L}(r^* = \text{d}_\text{cos}, \mathcal{D}) = \mathop{\mathbb{E}}_{x_i, x_j, \pi_\text{0} \in \mathcal{D}}[\text{log}(\sigma(r^*(x_i, \pi_\text{0})-r^*(x_j, \pi_\text{0})))]$
\State \text{Update network $f, g$ and $h$ to minimize} $\mathcal{L}_\textit{PrO}$
\EndFor
\State \textbf{return} encoder network $f$ and \textbf{discard} $g, h$ 
\end{algorithmic}
\end{algorithm}

% \section{Final copy}

% You must include your signed IEEE copyright release form when you submit your finished paper.
% We MUST have this form before your paper can be published in the proceedings.

% Please direct any questions to the production editor in charge of these proceedings at the IEEE Computer Society Press:
% \url{https://www.computer.org/about/contact}.

% \begin{figure*}\centering
% \subfloat[]{\label{a}\includegraphics[width=.245\linewidth]{pics/train-con-tnse.png}}\hfill
% \subfloat[]{\label{b}\includegraphics[width=.24\linewidth]{pics/test-con-tnse.png}}\hfill
% \subfloat[]{\label{c}\includegraphics[width=.248\linewidth]{pics/train-tsne.png}}\hfill%\par 
% \subfloat[]{\label{d}\includegraphics[width=.24\linewidth]{pics/test-tsne.png}}
% \caption{T-SNE visualization of latent space of (a) validation set after binary contrastive learning (b) training set and (c) test set after preference optimization}
% \label{fig}
% \end{figure*}
% \clearpage
% \setcounter{page}{1}
% \maketitlesupplementary
\begin{table*}[t]
\centering
\setlength{\tabcolsep}{8pt}
\caption{Multiclass classification results. SupCon-n denote n-classes supervised contrastive learning and SupCon-2 is the first stage of ConPro. ImageNet denotes a pre-trained model on ImageNet dataset. For MAEE score, the lower, the better}\label{tab1}
\begin{tabular}{lcccccc}
\hline
 &  \multicolumn{3}{c}{Papilledema} & \multicolumn{3}{c}{VinDr-Mammo}\\
\cmidrule(r){2-4} \cmidrule(r){5-7} 
{Methods}&Macro F1 & Recall &MAEE & Macro F1 &Recall & MAEE\\
\hline
% $\mathrlap{\textit{Binary classification task:}}$ \Tstrut \\
% SupCon-2  &  0 & 0& 0 $\pm$ 0 &  0 $\pm$ 0 & 0 $\pm$ 0& 0 $\pm$ 0\\
% SupCon-n  &  0 & 0& 0 $\pm$ 0 &  0 $\pm$ 0 & 0 $\pm$ 0& 0 $\pm$ 0\\
% SimCLR  &  0 & 0& 0 $\pm$ 0 &  0 $\pm$ 0 & 0 $\pm$ 0& 0 $\pm$ 0\\
% ImageNet  & 53.2  & 55.5 & 0 $\pm$ 0 &  0 $\pm$ 0 & 0 $\pm$ 0& 0 $\pm$ 0\\
% ConPrO (ours) & 67.4 & 67.9& 0 $\pm$ 0 &  0 $\pm$ 0 & 0 $\pm$ 0& 0 $\pm$ 0\\
% \hline
% $\mathrlap{\textit{Multiclass classification task:}}$ \Tstrut \\
% From Scratch  & 26.2  & 15.1 & 0 $\pm$ 0 &  0 $\pm$ 0 & 0 $\pm$ 0& 0 $\pm$ 0\\
ImageNet  & 38.7 $\pm$ 4.2 & 39.8 $\pm$ 3.1 & 6.4 $\pm$ 0.40 
&   17.9 $\pm$ 2.0 & 20.6 $\pm$ 0.9 & 2.9 $\pm$ 0.06 \\
SupCon-2  &  46.3 $\pm$ 5.5 & 47.4 $\pm$ 4.7 & 5.2 $\pm$ 0.30 
&  34.9 $\pm$ 0.7 & 33.6 $\pm$ 1.1 & 2.4 $\pm$ 0.03\\
SupCon-n  &  45.5 $\pm$ 4.1 & 46.9 $\pm$ 3.6 & 5.0 $\pm$ 0.26 
&  20.8 $\pm$ 1.5 & 23.1 $\pm$ 1.2 & 2.9 $\pm$ 0.03\\
SimCLR & 40.3 $\pm$ 3.9 & 43.8 $\pm$ 3.0  & {4.8 $\pm$ 0.26} 
&  25.1 $\pm$ 1.4 & 25.7 $\pm$ 1.2& 2.7 $\pm$ 0.05\\
ConPrO (ours) & \textbf{48.5 $\pm$ 3.8} & \textbf{49.4 $\pm$ 4.1} & \textbf{4.8 $\pm$ 0.25} 
&  \textbf{35.6 $\pm$ 0.8} & \textbf{34.7 $\pm$ 1.0} &  \textbf{2.4 $\pm$ 0.03}\\
\hline
\end{tabular}
\end{table*}

\section{Experiments}

\textbf{Datasets and Preprocessing.} We study two real-world datasets that contain severity labels: Papilledema and VinDr-Mammogram. 
% and Ocular Hypertension Treatment Study (OHTS)\footnote{OHTS was conducted retrospectively using human subject data made available in authorized access by National Eye Institute. No ethical approval was required for OHTS.}. 
The first two datasets contain discrete class labels while VinDr measures symptom severity on a continuous scale. Details of the statistics and preprocessing of each dataset are described below. 
\begin{itemize}
  % \item \textit{Ocular Hypertension Treatment Study} dataset, funded by the National Eye Institute, comes from a study conducted across multiple centers with a total of 1,636 subjects aged between 40 and 80 years. OHTS comprises over 74,000 fundus images from a prospective treatment trial aimed at evaluating whether treatment of increased intraocular pressure (IOP) reduces the risk for developing primary open-angle glaucoma (POAG). The severity of glaucoma is assessed by ophthalmologists using the mean deeviation (MD)-index, which indicates the average change in a patient's visual sensitivity across the portion of a visual field tested during a standardized assessment (30 degrees in the OHTS study) compared to an age-adjusted reference field of a healthy individual.
  \item \textit{Papilledema} is a controlled dataset comprising 331 pediatric fundus images obtained clinically from 105 subjects from 2011 to 2021.
  % at Children’s Hospital Los Angeles and the University of California, Los Angeles, from 2011 to 2021.
  The dataset contains a five-level severity rating for Papilledema. De-identified clinical datasets were uploaded to the HIPAA-compliant Research Electronic Data Capture (REDCap) database.
  % at the University of Southern California. 
  \item \textit{VinDr-Mammo} \cite{VindrMammo} is a public Vietnamese collection of full-field digital mammography comprising 5,000 four-view examinations with breast-level evaluations and annotated findings between 2018 and 2020. These examinations underwent independent double readings, with any disagreements resolved through third-party radiologist arbitration. The authors state that there are no ethical concerns. Approval was granted by the Institutional Review Boards of Hanoi Medical University Hospital and Hospital 108 to release de-identified data. The VinDr-Mammo dataset assesses the Breast Imaging-Reporting and Data System (BI-RADS) for breast level. It has 7 categories from 0 to 6 and be used as a risk evaluation and quality assurance tool. The datasets only contain the mammograms with BI-RADS from 1 to 5. In this work, we used this assessment to estimate the severity of the targeted breast. 
\end{itemize}

\noindent
\textbf{Evaluation.} We evaluate the final image representation using standard protocols \cite{simclr21, simmim20}. This involves training a linear classifier on the frozen-weight feature encoder, using the validation F1 score to choose the best model. That model is then used to compute the final scores on the test set. We evaluate the representation vector via classification tasks and use several evaluation metrics including Top-1 F1 scores (macro F1), Recall and Mean Absolute Exponential Error (MAEE). We proposed to used MAEE as a variant of Mean Absolute Error (MAE) for severity classification problems. MAEE is computed as
\begin{equation}
\text{MAEE} = \frac{1}{n}\sum_{i}^{n} \mathbf{e}^{|y_i-\hat{y}_i|}
\end{equation}
Both MAE and MAEE measure the error between predictions and true levels of severity.
However, while MAE evaluates regression problems with a linear penalty, MAEE assigns exponential penalties for incorrect severity level predictions. %Intuitively, as the wrong prediction ``go far'' from ground true, penalty weights are exponentially put on.

\noindent
\textbf{Experimental Setup.} We trained all datasets on a GTX 3090 with a batch size of 16. 
% Because a patient may take fundus/MRI photograph more than one time, we split our test/val and train by subjects's ids, not by images, to avoid data leakage. 
% Thus, the training dataset for Con and PrO are the same. 
Our choice of feature encoder is Resnet-50. % only, other encoder is not investigated in this paper. 
If not stated otherwise, our explorations utilize the following settings.
\begin{itemize}
    \item \textit{Data splitting:} Since each subject may have multiple visits or images have multiple views, we split train/val/test by $70/15/15$ for Papilledema dataset and $72/8/20$ for VinDr-Mammo by subject IDs to avoid data leakage. We chose pairs for preference optimization by randomly selecting $10^5$ pairs with replacements for training and $10^4$ for evaluation.
    \item \textit{Optimizer:} We use stochastic gradient descent (SGD) with momentum 0.9. We update the ResNet-50 encoder with a learning rate of $10^{-3}$ and the projection head with a learning rate of $0.01$ for both Supervised Contrastive Learning (Con) and Preference Comparison (PrO).
    \item \textit{Projection Head:} Resnet-50 outputs $2048$-d vectors. The projection head $g(\cdot)$ is a fully-connected (FC) layer with a 256-d output. In the `PrO' step, the projection head $h(\cdot)$ of preference comparison is an FC layer that keeps the preference vectors on the same dimension with normality.
    \item \textit{Prediction Head:} We use a 2-layer MLP with a hidden dimension of 256. The activation function is ReLU, and we use 10\% dropout. For fine-tuning, since both datasets are unbalanced, we use Cross Entropy with estimated class weights as the loss function.
\end{itemize}

\begin{figure*}\centering
\subfloat[SupCon-5 Train-set]{\label{fig:resulta}\includegraphics[width=.16\linewidth]{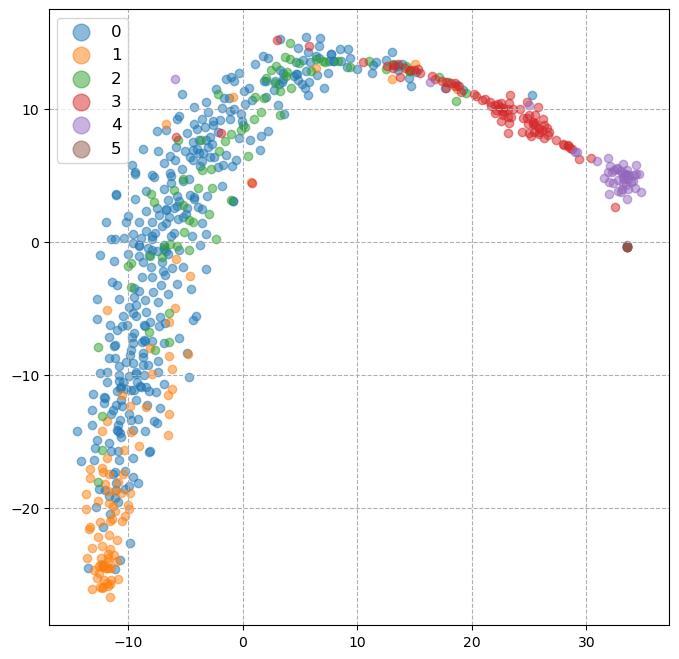}}\hfill
\subfloat[SupCon-2 Train-set]{\label{fig:resultb}\includegraphics[width=.16\linewidth]{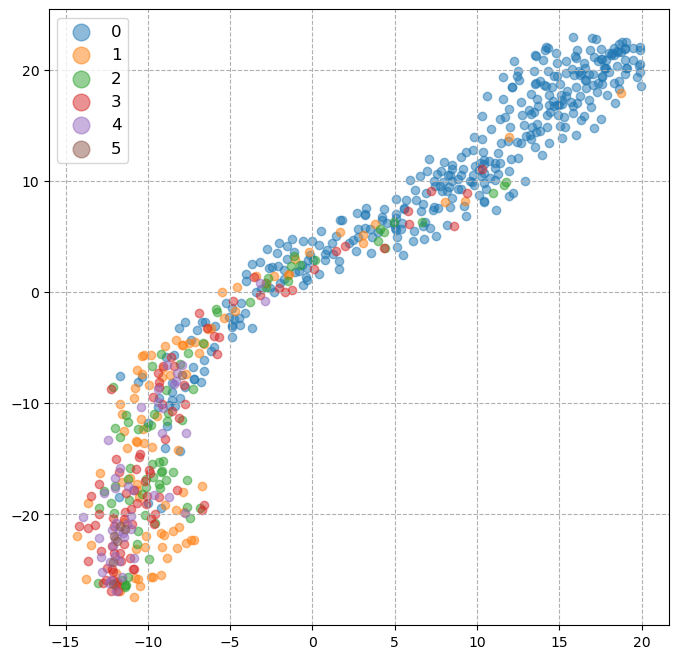}}\hfill
% \centering
\subfloat[ConPrO Train-set]{\label{fig:resultc}\includegraphics[width=.16\linewidth]{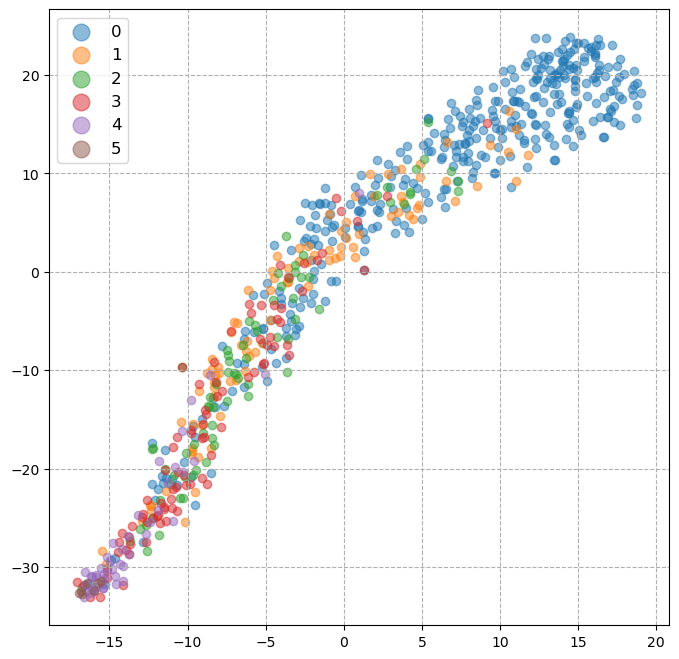}}\hfill%\par
\subfloat[SupCon-5 Test-set]{\label{fig:resultd}\includegraphics[width=.162\linewidth]{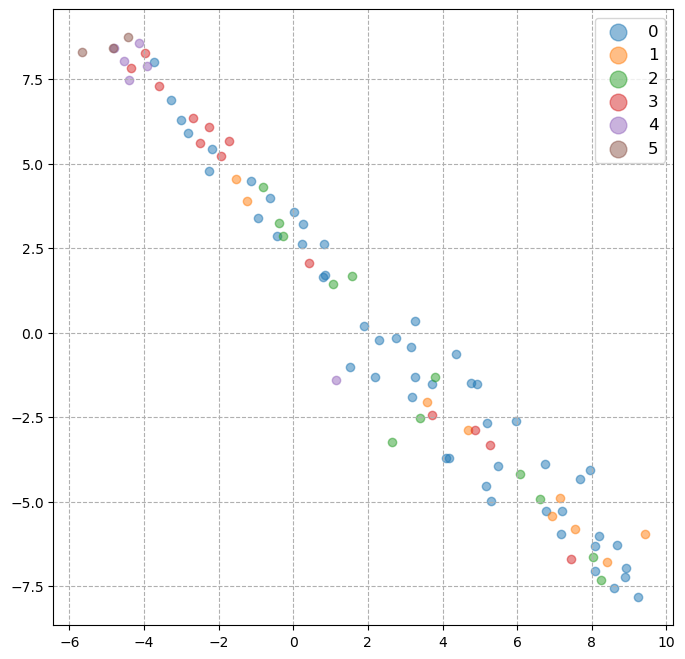}}\hfill
\subfloat[SupCon-2 Test-set]
{\label{fig:resulte}\includegraphics[width=.158\linewidth]{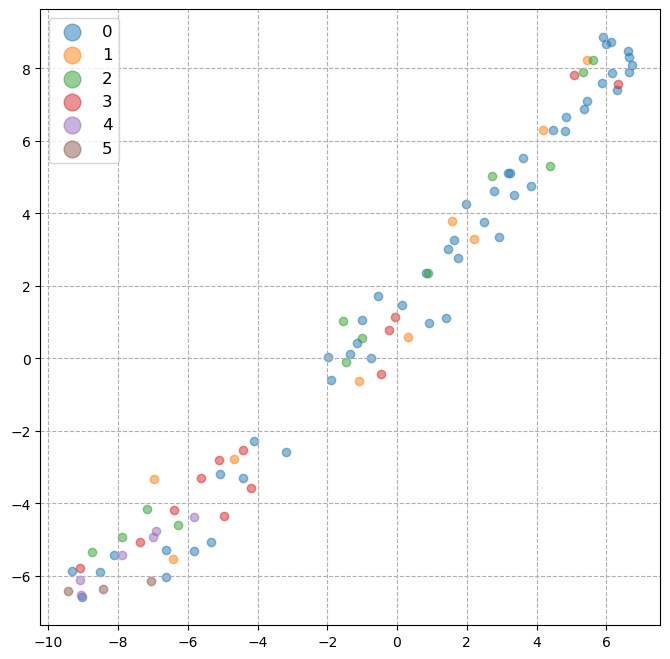}}\hfill
% \centering
\subfloat[ConPrO Test-set]{\label{fig:resultf}\includegraphics[width=.158\linewidth]{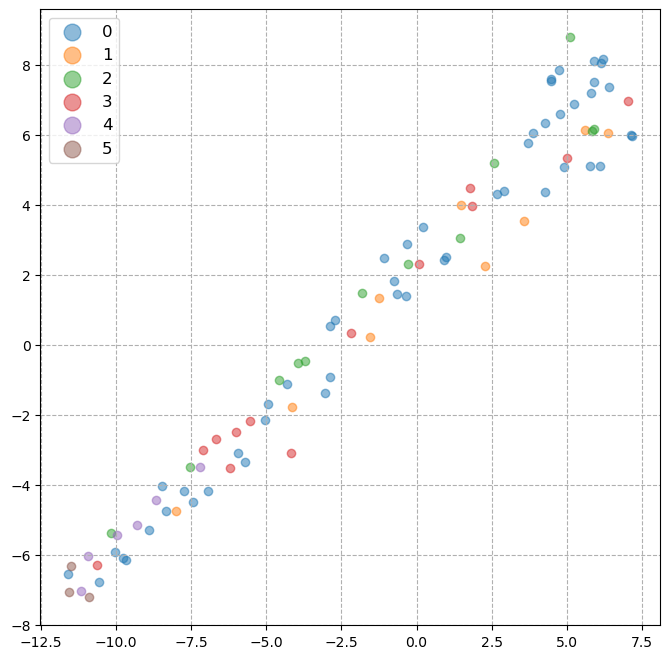}} 
\caption{T-SNE visualization of representation vectors of (a) training and (b) test set after supervised contrastive learning (c) training (d) test set after preference optimization. The plots are samples from the Papilledema dataset. All figures use cosine distance. Label '0' denotes normality, and ``1-5'' denotes increasing level of severity.}
\label{fig:result}
\end{figure*}

\begin{figure*}\centering
\subfloat[SupCon-5 Train-set]{\label{fig:mamresulta}\includegraphics[width=.16\linewidth]{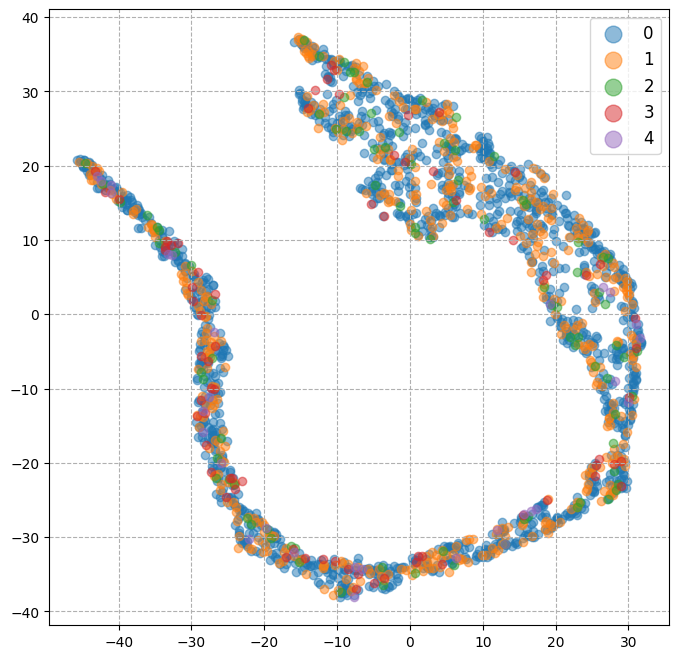}}\hfill
\subfloat[SupCon-2 Train-set]{\label{fig:mamresultb}\includegraphics[width=.16\linewidth]{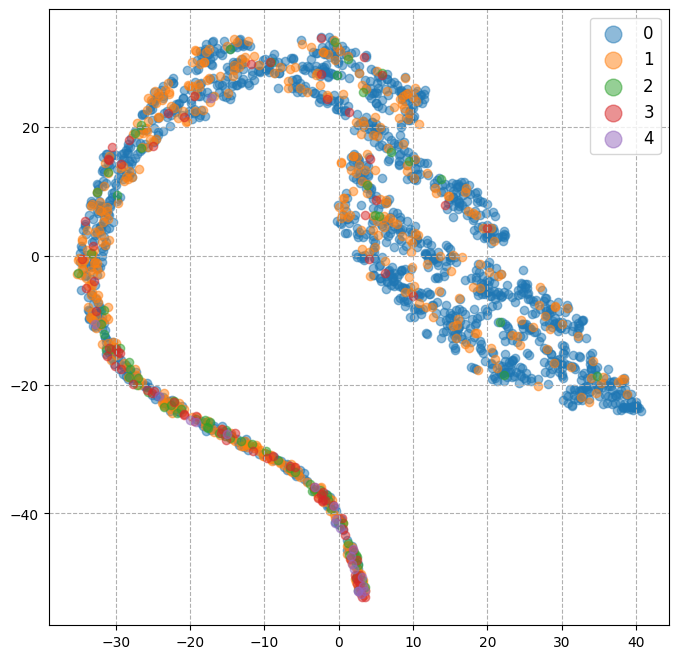}}\hfill
% \centering
\subfloat[ConPrO Train-set]{\label{fig:mamresultc}\includegraphics[width=.16\linewidth]{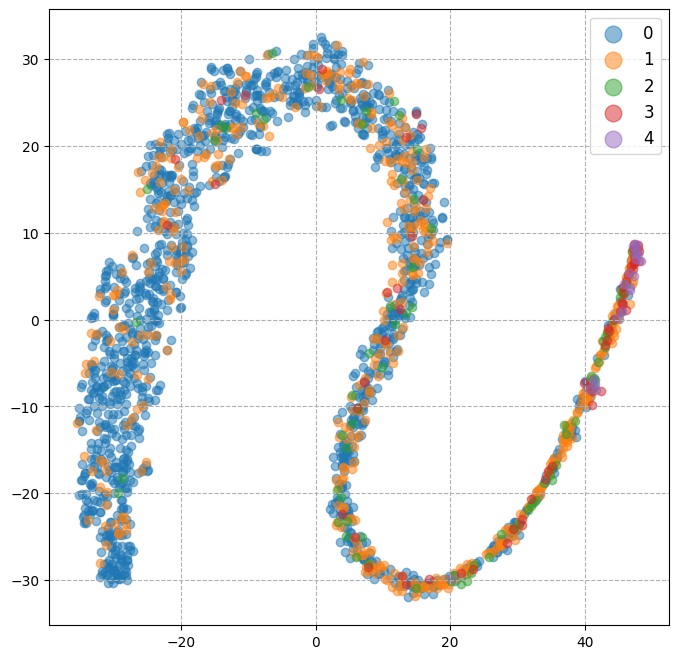}}\hfill%\par
\subfloat[SupCon-5 Test-set]{\label{fig:mamresultd}\includegraphics[width=.162\linewidth]{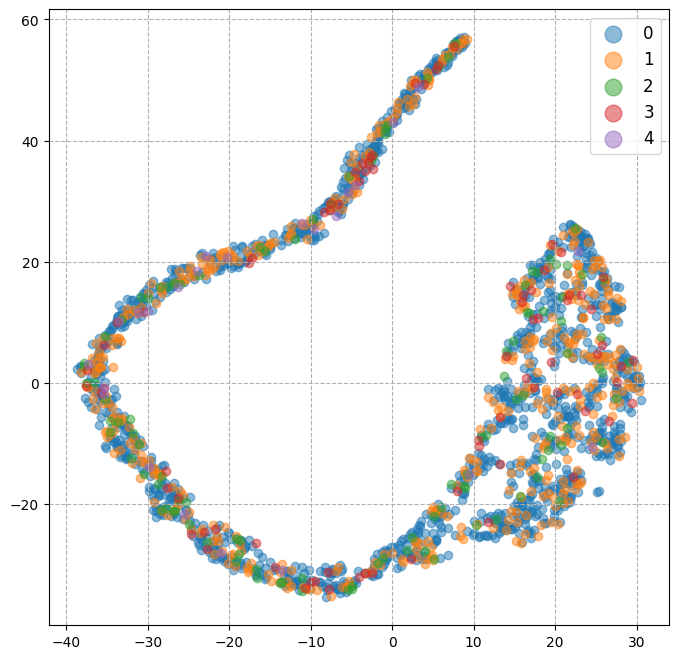}}\hfill
\subfloat[SupCon-2 Test-set]
{\label{fig:mamresulte}\includegraphics[width=.158\linewidth]{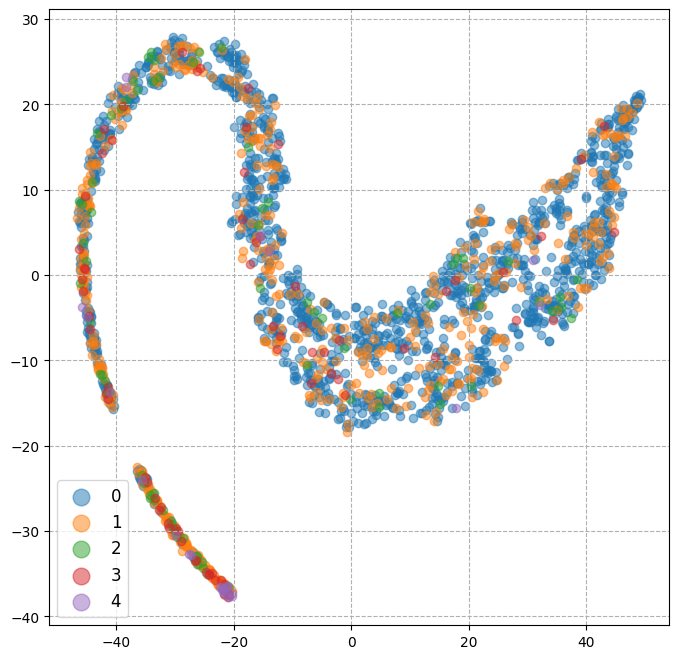}}\hfill
% \centering
\subfloat[ConPrO Test-set]{\label{fig:mamresultf}\includegraphics[width=.158\linewidth]{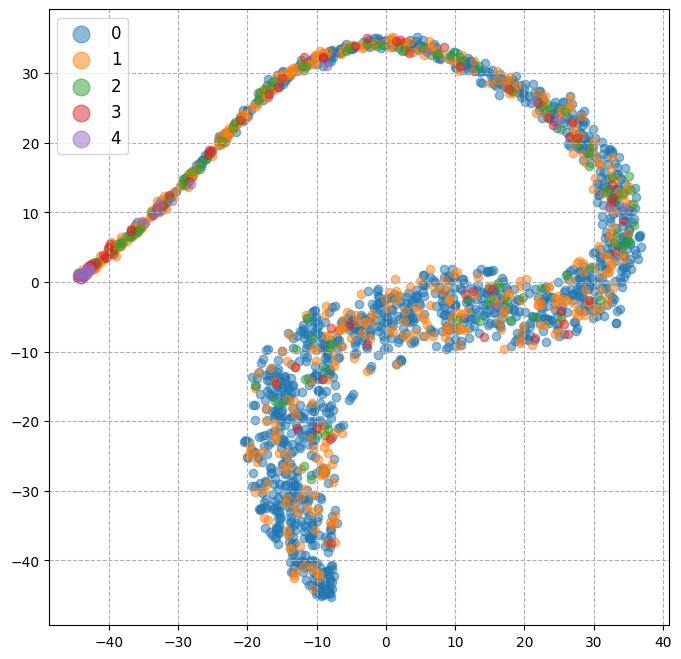}} 
\caption{T-SNE visualization of representation vectors of (a) training and (b) test set after supervised contrastive learning (c) training (d) test set after preference optimization. The plots are 2000 random samples from the VinDr-Mammo dataset. All figures use cosine distance. Label '0' denotes normality, and ``1-4'' denotes increasing level of severity.}
\label{fig:mamresult}
\end{figure*}

\section{Results}
\noindent
\textbf{Classification task performance.} The baseline is Resnet-50 pre-trained on the ImageNet dataset. Other SOTA methods include self-supervised (SimCLR) and supervised (SupCon-n) models. For all baselines we freeze the Resnet pre-trained weights and only fine-tune the prediction head on our target task. 
Table \ref{tab1} shows that ConPro outperforms the ImageNet baseline, SimCLR, and SupCon-n (SupCon-2 is the ``Con" step of our method) on all metrics in both datasets. ConPrO reaches the highest macro F1 score of 48.55\% and 35.6\%, respectively, in the Papilledema dataset on the 6-class classification task and the VinDr-Mammo dataset on the 5-class classification task. 
% It reaches the best-macro F1 score compared with all other methods. 
Compared to the ``Con'' step, the ``PrO'' step injects useful information from the ``Con'' step to latent space, boosting the performance by 4.8\% and 2\% on both datasets.
% Furthermore, we can see the trade-off of MAEE versus F1 score between SimCLR (self-supervised) and ConPro, SupCon-5 (supervised).

\noindent
\textbf{ConPrO better represents severity in feature space.} As shown in Fig. \ref{fig:result} and \ref{fig:mamresult}, we qualitatively evaluate the feature representations of our method on both datasets.
Comparing the `PrO' (Fig. \ref{fig:resultb} and \ref{fig:mamresultb}) step with the `Con' step (Fig. \ref{fig:resultc} and \ref{fig:mamresultc}), we show that the preference optimization successfully re-arranges the abnormal samples with respect to severity classes. The same behavior can be seen in the test set. Moreover, SupCon-5 (Fig. \ref{fig:resulta}) shows discrimination of severity classes, but the positions of the embeddings are not relative to the severity scores, which is not ideal for severity interpretation.

\begin{figure}\centering
\subfloat[MAE = 0.88 and MAEE = 5.05]{\label{fig:maeea}\includegraphics[width=.48\linewidth]{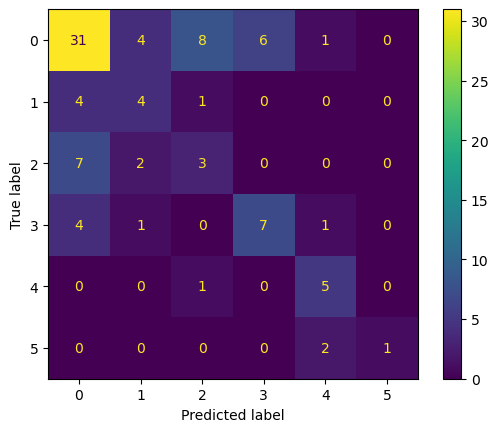}}\hfill
\subfloat[MAE = 0.86 and MAEE = 5.24]{\label{fig:maeeb}\includegraphics[width=.48\linewidth]{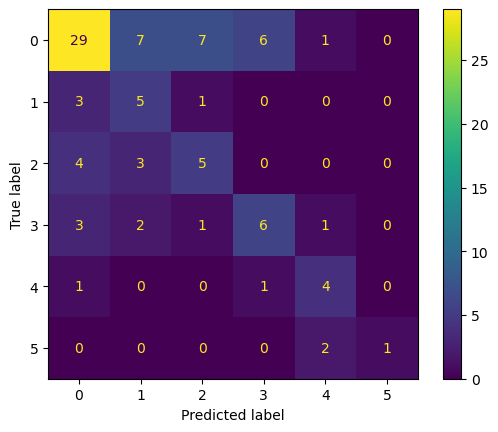}}\hfill%\par 
\caption{Confusion matrices of same setting on two independent run. Two matrix have the same F1 score but different in MAE and MAEE}
\label{fig:maeevsmae}
\end{figure}

\noindent
\textbf{MAE versus MAEE.}
Different from the F1 score that captures exact classification prediction, MAE and MAEE take prediction error into account. 
Fig. \ref{fig:maeevsmae} represents two confusion matrices having the same F1 score. While Fig. \ref{fig:maeeb} shows a better MAE score, Fig. \ref{fig:maeea} presents a greater value of MAEE. 
MAEE shows more sensitivity to incorrect predictions that deviate significantly from the ground truth, such as misclassifications between severity labels `3' and `4' as `0'.
This study opts to utilize MAEE as it helps us identify potential serious incorrect predictions in severity classification since there is no distinct boundary between severity classes. 
For instance, ophthalmologists usually group classes as mild $\{$1,2$\}$, moderate $\{$3$\}$, and severe $\{$4,5$\}$ since the finer annotation scale by experts also tends to have uncertainty. For mammogram examination, the BI-RADS score is also grouped into normal $\{$1$\}$, benign  $\{$2,3$\}$, and malignant $\{$4,5$\}$.

\noindent
\textbf{Choices of normality indicator.}
To calculate the reward function in equation~\ref{eq:proloss}, we need to compute the distance from each vector in pairs with respect to the normal class. Thus, ``normality'' is abstract and represented by a cluster of vectors. We randomly chose $n$ vectors in the normal class to get the mean of these vectors to get a single anchor representation for representing ``normality" (with respect to lack of severity). Table \ref{tab:metricvsnormalvecs} shows how F1 scores and MAEE change when varying the number of referenced vectors. Interestingly, by increasing the number of reference vectors, the framework lowers MAEE error across classes. As a trade-off, the macro F1 score will fall. 
% MAEE decrement indicates that the predictions do not go too far from ground truth on average.
The intuition is that by updating model parameters in the ``Pro'' step, we also update the normality anchor. By choosing the mean of multiple normal vectors as an anchor, we attempt to empirically lower the effect of model updating (no theoretical proof is derived in this work and is left for future work).
%Yet, this leaves a research question on how to choose refereed vectors for future research.
\begin{table}[h]
\centering
\caption{F1 Scores and MAEE versus the number of referenced vectors per pairs in the ``PrO'' step. For MAEE score, the lower, the better} \label{tab:metricvsnormalvecs}
\begin{tabular}{lccc}
% \hline
& \multicolumn{3}{c}{Number of reference vectors} \\
\cmidrule(r){2-4}
   Metric & 1 & 10 & 20 \\
\hline
  Macro F1 & \textbf{46.3 $\pm$ 5.7} & 45.2 $\pm$ 5.1 & 43.7 $\pm$ 6.3 \\
  MAEE & 5.4 $\pm$ 0.52 & 5.4 $\pm$ 0.42 & \textbf{5.2 $\pm$ 0.39}
\end{tabular}
\end{table}

\section{Discussion}

\noindent
\textbf{Explainable AI for Severity Ranking.}
Definitive reasons for the success of contrastive learning still remain incomplete in published literature. On the theoretical front, some \cite{Oord2018RepresentationLW} argue that the success is attributed to maximizing mutual information, while others emphasize the importance of the loss function \cite{Hjelm2018LearningDR, predictivecoding20}. However, in terms of explainability, it remains an open question what attributes contribute to positive pairs and what distinguishes negative pairs. 
The majority of literature \cite{Pillai_2022_CVPR, CLexplain21, Miller_2021, lin2023contrastive} uses contrastive (counterpart) methods to explain Deep Learning models. Yet, there are limited works \cite{explainseqrecCLs22, NEURIPS2020_4c2e5eaa} that focus on interpreting representations in contrastive learning. In this paper, we inject clinical condition severity knowledge into medical imaging representations, but we do not know what knowledge the latent space learned in deciding which sample is more severe. Thus this problem remains a topic for future investigation.

%Finally, the current trends in XAI is largely developer-centric which is driven by deep learning modeling advances in computer science. 
Subject matter expert-centric explanations, such as clinical judgments, may differ from what an AI model learns \cite{Expert_Informed_2022}. 
Performance of image-driven explainable AI (XAI) in clinical settings tends to degrade under three major pathological characteristics \cite{Saporta2022BenchmarkingSM}: multiple instances (pathology has multiple possible instantiations of interest and it is ranked variably by the preference of experts), size variety (instance size may vary between subjects, heterogeneity of clinical presentation, variability in severity between patients, and longitudinal changes in clinical manifestations that may be more important for diagnostic consideration than the severity of pathology at presentation), and pathology shape complexity. Reconciling user-centric and expert-centric explanations is a yet to be fully solved research problem wherein preference optimization can be advantageous.

% \noindent
% \textbf{Relation to Clustering}

\noindent
\textbf{Severity Indicator in Medical Domain.}
In interpreting ophthalmologic images, physicians frequently use comparisons in making diagnoses. 
For instance, in evaluating for glaucoma, asymmetry in the cup-to-disc ratio between two eyes of the same patient has predictive value for glaucoma diagnosis \cite{QIU20171229}.
Thus, it is easier for ophthalmologists to compare two fundus images to decide which one has more asymmetry, rather than assign a class label for each image. 
For that reason, preference comparison may play an important part in improving diagnosis. This also bring up multiple challenge including 
\begin{itemize}
    \item \textit{Severity indicators on multiple perspective of diagnosis:} In glaucoma diagnosis, assessments often rely on either fundus photos or results from visual field tests. It's crucial to recognize that the severity reflect from fundus photo different from severity of visual field test although there may be a strong correlation. There is often discrepancy between metrics of structure (photos) and function (visual field tests) in ophthalmology (and other fields of medicine). Some patients with glaucoma have normal visual fields (pre-perimetric glaucoma). 
    \item \textit{Severity preference on multiple pathologies:} One image may endure multiple conditions (e.g. VinDr-Mammo dataset represent 15 types of pathologies and each image may contain more than one type). The challenge lies in comparing and prioritizing the severity of multiple pathologies within the same image.
    % \item \textit{Experts' preference on comparison:} 
\end{itemize}

% \noindent
% \textbf{Severity Preference on multiple Pathologies} 
% Furthermore, one image may contain multiple pathologies (ex. Vin-Dr Mammo dataset represent 15 type of Pathology). Literature pays attention to developer models that can discriminate these pathologies. However, one subject may endure multiple pathologies not just one, it is reasonable to evaluate 

% \noindent
% \textbf{Experts' consensus} 
% It is not always the case that all experts have the same perspective of what is going on on medical images. Several works [?] have tried to meet the consensus between multiple experts in severity classification.

\section{Conclusion}

This paper presents a representation learning method to inject severity information in the latent representation space. We meticulously examine the components of the suggested framework and show that (1) ConPrO not only demonstrates a dependable representation via TSNE visualization but also surpasses state-of-the-art algorithms in classification tasks by at least 6\% in F1 score, (2) proposed MAEE metric penalizes serious incorrect prediction which fit well to the severity classification problem, (3) choosing good ``normality'' anchor can help reduce MAEE score.
Finally, we discuss several problems of preference comparison and explainable AI as potential directions for future work.
{
    \small
    \bibliographystyle{ieeenat_fullname}
    \bibliography{main}
}


\end{document}